\documentclass{article}
\usepackage{ijcai23}

\usepackage{booktabs,multirow}

\usepackage{times}
\usepackage{soul}
\usepackage{url}
\usepackage[hidelinks]{hyperref}
\usepackage[utf8]{inputenc}
\usepackage[small]{caption}
\usepackage{subcaption}
\usepackage{graphicx}
\usepackage{amsmath,amsfonts,amssymb,amsthm}
\usepackage{mathtools,thmtools,thm-restate}
\usepackage{booktabs}
\usepackage{algorithm}
\usepackage[noend]{algpseudocode}
\usepackage{newfloat}
\usepackage{listings}
\usepackage{comment}
\usepackage[dvipsnames]{xcolor}
\usepackage{array}
\usepackage[noabbrev,nameinlink,capitalize]{cleveref}
\usepackage{titlesec}
\usepackage{enumitem}
\usepackage{etoolbox}
\usepackage{xspace}
\usepackage{setspace}
\usepackage{relsize}
\usepackage{stmaryrd}
\usepackage[mode=text]{siunitx}
\usepackage{nicefrac}       

\usepackage{tikz}
\usepackage{pgf}
\usepackage{forest}

\usetikzlibrary{arrows,decorations.pathmorphing,decorations.footprints,fadings,calc,trees,mindmap,shadows,decorations.text,patterns,positioning,shapes,matrix,fit}
\usetikzlibrary{graphs}
\usetikzlibrary{decorations.pathreplacing}
\usetikzlibrary{arrows,shadows,backgrounds}
\usetikzlibrary{arrows.meta}
\usetikzlibrary{positioning,fit}
\usetikzlibrary{automata}
\usetikzlibrary{matrix}
\usetikzlibrary{karnaugh}
\usetikzlibrary{shapes.symbols,shapes.misc,shapes.arrows}
\usetikzlibrary{matrix,chains,scopes,decorations.pathmorphing}

\usetikzlibrary{fit,calc}

\definecolor{darkslategray}{rgb}{0.18, 0.31, 0.31} 
\definecolor{platinum}{rgb}{0.9, 0.89, 0.89} 
\definecolor{gray}{rgb}{.4,.4,.4}
\definecolor{midgrey}{rgb}{0.5,0.5,0.5}
\definecolor{middarkgrey}{rgb}{0.35,0.35,0.35}
\definecolor{darkgrey}{rgb}{0.3,0.3,0.3}
\definecolor{darkred}{rgb}{0.7,0.1,0.1}
\definecolor{midblue}{rgb}{0.2,0.2,0.7}
\definecolor{darkblue}{rgb}{0.1,0.1,0.5}
\definecolor{darkgreen}{rgb}{0.1,0.5,0.1}
\definecolor{defseagreen}{cmyk}{0.69,0,0.50,0}

\newtheorem{proposition}{Proposition}

\newtheorem{example}{Example}

\crefname{theorem}{Theorem}{Theorems}
\crefname{lemma}{Lemma}{Lemmas}
\crefname{proposition}{Proposition}{Propositions}
\crefname{definition}{Definition}{Definitions}
\crefname{corollary}{Corollary}{Corollaries}
\crefname{example}{Example}{Examples}
\crefname{claim}{Claim}{Claims}
\crefname{assumption}{Assumption}{Assumptions}
\crefname{enumi}{}{}

\newcommand{\fml}[1]{{\mathcal{#1}}}

\newcommand{\tn}[1]{\textnormal{#1}}
\newcommand{\tbf}[1]{\textbf{#1}}
\newcommand{\mbf}[1]{\ensuremath\mathbf{#1}}
\newcommand{\msf}[1]{\ensuremath\mathsf{#1}}
\newcommand{\mbb}[1]{\ensuremath\mathbb{#1}}

\newcommand{\waxp}{\ensuremath\mathsf{WAXp}}
\newcommand{\wcxp}{\ensuremath\mathsf{WCXp}}
\newcommand{\axp}{\ensuremath\mathsf{AXp}}
\newcommand{\cxp}{\ensuremath\mathsf{CXp}}

\newcommand{\bigland}{\ensuremath\bigwedge}

\newcounter{tableeqn}[table]

\DeclareMathOperator*{\limply}{\rightarrow}

\newcommand{\jnoteF}[1]{}

\newcolumntype{L}[1]{>{\raggedright\let\newline\\\arraybackslash\hspace{0pt}}m{#1}}
\newcolumntype{C}[1]{>{\centering\let\newline\\\arraybackslash\hspace{0pt}}m{#1}}
\newcolumntype{R}[1]{>{\raggedleft\let\newline\\\arraybackslash\hspace{0pt}}m{#1}}

\tikzset{
  0 my edge/.style={densely dashed, my edge},
  my edge/.style={-{Stealth[]}},
}

\def\HiLi{\leavevmode\rlap{\hbox to \linewidth{\color{platinum}\leaders\hrule height .8\baselineskip depth .5ex\hfill}}}

\setitemize{noitemsep,topsep=0pt,parsep=0pt,partopsep=0pt}
\setenumerate{noitemsep,topsep=0pt,parsep=0pt,partopsep=0pt}
\setlist{nosep,leftmargin=0.45cm}

\providetoggle{long}
\settoggle{long}{false}

\makeatletter
\algnewcommand{\LineComment}[1]{\Statex \hskip\ALG@thistlm \(\triangleright\) #1}
\makeatother

\begin{document}

%

\title{Delivering Inflated Explanations}

\iftrue
%

\author{
Yacine Izza$^1$
\and
Alexey Ignatiev$^2$
\and
Peter Stuckey$^2$
\and
Joao Marques-Silva$^{3}$
\affiliations
$^1$CREATE, National University of Singapore, Singapore\\
$^2$Monash University, Melbourne, Australia\\
$^3$IRIT, CNRS, Toulouse, France\\
\emails
izza@com.nus.edu.sg,
alexey.ignatiev@monash.edu,
peter.stuckey@monash.edu,
joao.marques-silva@irit.fr
}
\fi

\maketitle

\begin{abstract}
In the quest 
for Explainable Artificial Intelligence (XAI) one of the questions 
that frequently arises given a decision made by an AI system is,
``why was the decision made in this way?''
Formal approaches 
to explainability build a formal model of the AI system and
use this to reason about the properties of the system.
Given a set of feature values for an instance to be explained, 
and a resulting decision, a formal 
\emph{abductive explanation} is a set of features, such that if they take
the given value will always lead to the same decision. 
This explanation is useful, it shows that only some features were used in
making the final decision. But it is narrow, it only shows that if the
selected features take their given values the decision is unchanged. Its
possible that some features may change values and still lead to the same
decision. In this paper we formally define \emph{inflated explanations}
which is a set of features, and for each feature of set of values 
(always including the value of the instance being explained),
such that the decision will remain unchanged. 
Inflated explanations are more informative than abductive explanations
since e.g 
they allow us to see if the exact value of a feature is important, or
it could be any nearby value.
Overall they allow us to better 
understand the role of each feature in the decision. 
We show that we can compute inflated explanations for not that much greater
cost than abductive explanations, and 
that we can extend duality results for
abductive explanations also to inflated explanations. 

\end{abstract}


\section{Introduction} \label{sec:intro}

The purpose of eXplainable AI (XAI) is to help human decision makers
in understanding the decisions made by AI systems.
It is generally accepted that XAI is fundamental to deliver
trustworthy AI~\cite{ignatiev-ijcai20,msi-aaai22}. In addition,
explainability is also at the core of recent proposals for the
verification of Artificial Intelligence (AI)
systems~\cite{seshia-cacm22}.
Nevertheless, most of the work on XAI offers no formal guarantees of
rigor (and so will be referred to as non-formal XAI in this paper).
Examples of non-formal XAI include model-agnostic
methods~\cite{guestrin-kdd16,lundberg-nips17,guestrin-aaai18},
heuristic learning of saliency maps (and their
variants)~\cite{muller-plosone15,muller-xai19-ch01,xai-bk19,muller-ieee-proc21},
but also proposals of intrinsic
interpretability~\cite{rudin-naturemi19,molnar-bk20,rudin-ss22}.
In recent years, comprehensive evidence has been gathered that attests
to to the lack of rigor of these (non-formal) XAI
approaches~\cite{ignatiev-ijcai20,iims-jair22}.

The alternative to non-formal explainability is \emph{formal XAI}.
Formal XAI proposes definitions of explanations, and algorithms for
their computation, that ensure the rigor of computed
explanations~\cite{darwiche-ijcai18,inms-aaai19,msi-aaai22}.
Despite its promise, formal XAI also exhibits a number of important
limitations, which include lack of scalability for some ML models, and
the computation of explanations which human decision makers may fail
to relate with.
This paper targets mechanisms for improving the clarity of computed
explanations.

An abductive explanation~\cite{darwiche-ijcai18,inms-aaai19} is a
subset-minimal set of features which correspond to a rule of the form:
if a conjunction of literals is true, then the prediction is the
expected one.
The literals associated with that rule are of the form $x_i=v_i$,
i.e.\ a feature is tested for a concrete value in its domain.
For categorical features with large domains, and more importantly for
real-value features, specifying a literal that tests a single value
will provide little insight.
For example, in the case of an ordinal feature, stating that the
height of a patient is 1.825m is less insightful that stating that the
height of a patient is between 1.75m and 1.84m.
Similarly, in the case of a categorical feature, an explanation that
indicates that the vehicle color is one of
$\{\tn{Red},\tn{Blue},\tn{Green},\tn{Silver}\}$ (e.g.\ with colors
$\{\tn{Black},\tn{White}\}$ excluded), is more insightful that stating
that the color must be $\tn{Blue}$.
For an explanations involving several features, the use of more
expressive literals will allow a human decision maker to relate the
explanation with several different instances.

This paper proposes \emph{inflated explanations}. In an inflated
explanation the literals of the form $x_i=v_i$ are replaced by
literals of the form $x_i\in\mbb{E}_i$, where $\mbb{E}_i$ is a subset
of the feature's domain. Furthermore, each $\mbb{E}_i$ is maximally
large, i.e.\ no proper superset of $\mbb{E}_i$ guarantees sufficiency
for the prediction.
%
Inflated explanations can be related with previous
works~\cite{darwiche-corr20,iims-jair22,darwiche-corr23}, but also
with earlier work on minimum satisfying
assignments~\cite{dillig-cav12}.
%
For example, recent work~\cite{iims-jair22} proposed algorithms for
computing explanations in the case of decision trees that report path
explanations, where the literals are taken from a concrete tree path;
however, no algorithm for inflating explanations is described.
In contrast, our contribution is to propose algorithms for
transforming abductive explanations into inflated (abductive)
explanations, and such that these algorithms are shown to be efficient
in practice.
Furthermore, we prove that contrastive explanations can also be
inflated, and that there exist different minimal hitting set
duality~\cite{inams-aiia20} relationships between inflated
explanations. The duality results regarding contrastive explanations
attest to the robustness of minimal hitting set duality in
explainability. 
The experimental results illustrate that in practice features can in
general be inflated, with some inflated to almost their original
domain.

The paper is organized as follows.
\cref{sec:prelim} introduces the notation and definitions used
throughout. This section also motivates the importance of inflating
explanations.
\cref{sec:iaxp} defines inflated  abductive explanations, and details
algorithms for their computation. Different algorithms are devised for
categorical and ordinal features.
Moreover,~\cref{sec:icxp} proposes two definitions for inflated
contrastive explanations and demonstrates that minimal hitting set
duality with respect to inflated abductive explanations in both
cases.
\cref{sec:res} demonstrates the practical interest of inflated
explanations.
\cref{sec:conc} concludes the paper.

\jnoteF{ToDo:\\
  The challenge: how to make explanations more expressive?\\
  Approach: we build on earlier
  work~\cite{darwiche-corr20,iims-jair22} proposing literals of
  explanations to be made more general. However, we propose algorithms
  that find maximal literals both for categorical and ordinal
  features.\\
  The computational problem we address is: \emph{Given some AXp, how
  to make the literals of the associated rule maximally expressive?}\\
  We also study the related problem of \emph{strengthening} the
  conditions for CXp's.
}

\section{Preliminaries} \label{sec:prelim}

\paragraph{Classification problems.}
%
We consider a classification problem, characterized by a set of
features $\fml{F}=\{1,\ldots,m\}$, and by a set of classes
$\fml{K}=\{c_1,\ldots,c_K\}$.
Each feature $j\in\fml{F}$ is characterized by a domain $\mbb{D}_j$.
As a result, feature space is defined as
$\mbb{F}={\mbb{D}_1}\times{\mbb{D}_2}\times\ldots\times{\mbb{D}_m}$.
A specific point in feature space represented by
$\mbf{v}=(v_1,\ldots,v_m)$
%
denotes an \emph{instance} (or an
\emph{example}).
Also, we use $\mbf{x}=(x_1,\ldots,x_m)$ to denote an arbitrary
point in feature space.
In general, when referring to the value of a feature $j\in\fml{F}$, we
will use a variable $x_j$, with $x_j$ taking values from $\mbb{D}_j$.
We consider two types of features $j \in \fml{F}$:
\emph{categorical features} where $\mbb{D}_j$
is a finite unordered set, and
\emph{ordinal features} where $\mbb{D}_j$ is a possibly infinite
ordered set. For ordinal (real-valued) features $j$
we use range notation $[\lambda(j),\mu(j)]$ to indicate the set of
values $\{ d ~|~ d \in \mbb{D}_j, \lambda(j) \leq d \leq \mu(j) \}$,
and  $[\lambda(j),\mu(j))$ to indicate the (half-open) set of values
$\{ d ~|~ d \in \mbb{D}_j, \lambda(j) \leq d < \mu(j) \}$.
%
%
A classifier implements a \emph{total classification function}
$\kappa:\mbb{F}\to\fml{K}$.
%
%
%
%
%
%
%
For technical reasons, we also require $\kappa$ not to be a constant
function, i.e.\ there exists at least two points in feature space with
differing predictions.

%
%
Given the above, we represent a classifier $\fml{M}$ by a tuple
$\fml{M}=(\fml{F},\mbb{F},\fml{K},\kappa)$.
%
%
Moreover, and given a concrete instance $(\mbf{v},c)$, an explanation
problem is represented by a tuple $(\fml{M},(\mbf{v},c))$.

\jnoteF{Classifiers used: monotonic classifiers \& tree ensembles.}

Throughout the paper, we consider the following families of
classifiers: monotonic classifiers, decision lists and tree
ensembles. These families of classifiers are well-known, and have been
investigated in the context of logic-based
explainability~\cite{msgcin-icml21,ims-sat21,ignatiev-ijcai20}.

\jnoteF{For ordinal features: $\mbb{D}_j=[\lambda(j),\mu(j)]$.\\
  Suggestion: intervals of the form $[a,b]$, with the exception
  of unbounded intervals; otherwise, the algorithms become cumbersome.}

\paragraph{Running examples.}
%
The monotonic classifier of~\cref{ex:runex01} will be used throughout
the paper.

\begin{example}[Running example -- ordinal features] \label{ex:runex01}
  We consider a monotonic classifier defined on ordinal features,
  adapted from~\cite[Example~1]{msgcin-icml21}.
  The classifier serves for predicting student grades. It is assumed
  that the classifier has learned the following formula (after being
  trained with grades of students from different cohorts):
  \[
  \begin{array}{rcl}
    S & = & \max\left[0.3 \times{Q}+0.6\times{X}+0.1\times{H},R\right]\\[2pt]
    M & = & \tn{ite}(S\ge9,A,\tn{ite}(S\ge7,B,\tn{ite}(S\ge5,C,\\[1.5pt]
    & & \tn{ite}(S\ge4,D,\tn{ite}(S\ge2,E,F)))))\\
  \end{array}
  \]
  The features represent the different components of assessment,
  namely $S$, $Q$, $X$, $H$ and $R$ denote, respectively, the final
  score, the marks on the quiz, the exam, the homework, and the mark
  of an optional research project. Each mark ranges from 0 to 10.
  (For the optional mark $R$, the final mark is 0 if the student opts
  out.)  The final score is the largest of the two marks, as shown
  above.
  The student's final grade $M$ is defined using an $\tn{ite}$
  (if-then-else) operator, and ranges from $A$ to $F$.
  Features $Q$, $X$, $H$ and $R$ are respectively numbered 1, 2, 3 and
  4, and so $\fml{F}=\{1,2,3,4\}$.
  Each feature takes values from $[0,10]$, i.e.\ $\lambda(i)=0$ and
  $\mu(i)=10$. The set of classes is $\fml{K}=\{A,B,C,D,E,F\}$, with
  ${F}\prec{E}\prec{D}\prec{C}\prec{B}\prec{A}$.
  Clearly, the complete classifier (that given the different marks
  computes a final grade) is monotonic.
  Moreover, and in contrast with~\cite{msgcin-icml21}, we will we
  consider the following point in feature space representing a
  student $s_1$, $(Q,X,H,R)=(5,10,5,8)$, with a predicted grade of
  $B$, i.e.\ $\kappa(5,10,5,8)=B$, given that $S=8$. 
  Moreover, and unless stated otherwise, the order in which features
  are analyzed throughout the paper will be $\langle1,2,3,4\rangle$.
\end{example}

The decision list of~\cref{ex:runex02} will also be used throughout
the paper.

\begin{example}[Running example -- categorical features] \label{ex:runex02}
  We also consider a decision list with categorical features.
  The classification problem is to assess risk accident, given two
  features, age segment (represented by variable $A$), and car color
  (represented by variable $C$).
  Let $\fml{F}=\{1,2\}$, $\fml{K}=\{0,1\}$
  $\mbb{D}_1=\{\tn{Adult},\tn{Young~Adult},\tn{Senior}\}$ (but where
  we use $\tn{Junior}$ instead of $\tn{Young~Adult}$)
  $\mbb{D}_2=\{\tn{Red},\tn{Blue},\tn{Green},\tn{Silver},\tn{Black},\tn{White}\}$.
  Let the decision list be,
  \[
  \begin{array}{llcc}
    \tn{IF} & A=\tn{Adult} & \tn{THEN} & \kappa(\mbf{x})=0 \\
    \tn{ELSE IF} & C=\tn{Red} & \tn{THEN} & \kappa(\mbf{x})=1 \\
    \tn{ELSE IF} & C=\tn{Blue} & \tn{THEN} & \kappa(\mbf{x})=1 \\
    \tn{ELSE IF} & C=\tn{Green} & \tn{THEN} & \kappa(\mbf{x})=1 \\
    \tn{ELSE IF} & C=\tn{Black} & \tn{THEN} & \kappa(\mbf{x})=1 \\
    \tn{ELSE} & & & \kappa(\mbf{x})=0 \\
  \end{array}
  \]
  Moreover, we consider the instance
  $(\mbf{v},c)=((\tn{Junior},\tn{Red}),1)$, i.e.\ a young adult with a
  red colored car represents a risk of accident.
\end{example}

\paragraph{Logic-based explainability.}
%
Two types of formal explanations have been studied:
abductive~\cite{darwiche-ijcai18,inms-aaai19} and
contrastive~\cite{miller-aij19,inams-aiia20}. Abductive explanations
broadly answer a \tbf{Why} question, i.e.\ \emph{Why the prediction?},
whereas contrastive explanations broadly answer a \tbf{Why Not}
question, i.e.\ \emph{Why not some other prediction?}.

Given an explanation problem, an abductive explanation (AXp) is a
subset-minimal set of features $\fml{X}\subseteq\fml{F}$ which, if
assigned the values dictated by the instance $(\mbf{v},c)$, are
sufficient for the prediction. This is stated as follows, for a chosen
set $\fml{X}$:
\begin{equation} \label{eq:axp}
  \forall(\mbf{x}\in\mbb{F}).\left[\bigland_{i\in\fml{X}}(x_i=v_i)\right]\limply(\kappa(\mbf{x})=c) 
\end{equation}
Monotonicity of entailment implies that there exist algorithms for
computing a subset-minimal set $\fml{X}\subseteq\fml{F}$ that are
polynomial on the time to decide~\eqref{eq:axp}~\cite{msi-aaai22}.

An AXp $\fml{X}$ can be interpreted as a logic rule, of the form:
\begin{equation} \label{eq:axprule}
  \tn{IF}~~\left[\bigland_{i\in\fml{X}}(x_i=v_i)\right]~~\tn{THEN}~~[\kappa(\mbf{x})=c] 
\end{equation}

Moreover, and given an explanation problem, a contrastive explanation
(CXp) is a subset-minimal set of features $\fml{Y}\subseteq\fml{F}$
which, if the features in $\fml{F}\setminus\fml{Y}$ are assigned the
values dictated by the instance $(\mbf{v},c)$, then there is an
assignment to the features in $\fml{Y}$ that changes the prediction.
This is stated as follows, for a chosen set $\fml{Y}\subseteq\fml{F}$:
\begin{equation} \label{eq:cxp}
  \exists(\mbf{x}\in\mbb{F}).\left[\bigland_{i\in\fml{X}\setminus\fml{Y}}(x_i=v_i)\right]\land(\kappa(\mbf{x})\not=c) 
\end{equation}
Similarly to the case of AXp's, monotonicity of entailment ensures
that there exist algorithms for computing a subset-minimal set
$\fml{Y}\subseteq\fml{F}$ that are polynomial on the time to
decide~\eqref{eq:cxp}~\cite{msi-aaai22}.

Following standard notation~\cite{msi-aaai22}, we use the predicate
$\waxp$ (resp.~$\wcxp$) to hold true for any set
$\fml{X}\subseteq\fml{F}$ for which~\eqref{eq:axp}
(resp.~\eqref{eq:cxp}) holds, and the predicate $\axp$ (resp.~$\cxp$)
to hold true for any subset-minimal (or  irreducible) set
$\fml{X}\subseteq\fml{F}$ for which~\eqref{eq:axp}
(resp.~\eqref{eq:cxp}) holds.

AXp's and CXp's respect a minimal-hitting set (MHS) duality
relationship~\cite{inams-aiia20}. Concretely, each AXp is an MHS of
the CXp's and each CXp is an MHS of the AXp's. MHS duality is a
stepping stone for the enumeration of explanations.

\jnoteF{AXp's, CXp's, duality.}

\begin{example} \label{ex:axp01}
  For the monotonic classifier of~\cref{ex:runex01}, and instance
  $((5,10,5,8),B)$, we can use existing
  algorithms~\cite{msgcin-icml21,cms-cp21,cms-aij23}, for computing
  one AXp. We use the proposed order for analyzing features:
  $\langle1,2,3,4\rangle$.
  we conclude that $Q$ must be included in the AXp, since increasing
  the value of $Q$ to 10 would change the prediciton.
  In contrast, given the value of $R$, and because we have fixed $Q$,
  then both $X$ and $H$ are dropped from the AXp.
  Moreover, we conclude that $R$ must be included in the AXp;
  otherwise, we would  be able to change the prediction to $A$ by
  increasing $R$.
  %
  As a result, the computed AXp is $\fml{X}=\{1,4\}$.
  (For a different order of the features, a different AXp would be
  obtained.)
  With this AXp, we associate the following rule:
  \[
  \tn{IF}~[{Q}=5\land{R}=8]~\tn{THEN}~[\kappa(\mbf{x})=B]
  \]
  %
\end{example}

\begin{example}  \label{ex:axp02}
  For the decision list running example (see~\cref{ex:runex02}), and
  given the instance $(\mbf{v},c)=((\tn{Junior}, \tn{Red}), 1)$,
  an AXp is $\{1,2\}$, meaning that,
  \[
  \tn{IF} ~~ A=\tn{Junior} \land C=\tn{Red} ~~ \tn{THEN} ~~
  \kappa(\mbf{x})=1
  \]
  where $\mbf{x}=(A,C)$.
\end{example}

Logic-based explainability is covered in a number of recent works.
Explainability queries are
studied in~\cite{marquis-kr20,marquis-kr21,hiims-kr21,marquis-dke22}.
%
Probabilistic explanations are investigated
in~\cite{kutyniok-jair21,barcelo-nips22}.
There exist proposals to account for constraints on the
inputs~\cite{rubin-aaai22}. 
%
A distinction between constrastive and counterfactual explanations is
studied in~\cite{lorini-jlc23}.
Additional recent works
include~\cite{mazure-sum20,kwiatkowska-ijcai21,mazure-cikm21,lorini-wollic22,darwiche-jlli23,amgoud-ijar23,katz-tacas23}.
In addition, there exist recent surveys summarizing the progress
observed in formal XAI~\cite{msi-aaai22,ms-corr22}. 

\paragraph{Motivating examples \& related work}
%
For the two running examples, let us anticipate what we expect to
obtain with inflated explanations.

\begin{example}[Explanations with more expressive literals -- ordinal features.]
  \label{ex:infxp01}
  For the monotonic classifier of~\cref{ex:runex01}, and instance
  $((5,10,5,7),B)$, one AXp is $\fml{X}=\{1,4\}$.
  Our goal is 
  to identify more general literals
  than the equality relational operator. Similarly to recent
  work~\cite{iims-jair22}, we use the set-membership ($\in$)
  operator.
  The purpose of the paper is to propose approaches for obtaining more
  expressive rules using the $\in$ operator.
  As a result, instead of the rule from~\cref{ex:axp01}, a more
  expressive rule would be,
  \[
  \tn{IF}~[{Q}\in[0,6.6]\land{R}\in[7,8.8]]~\tn{THEN}~[\kappa(\mbf{x})=B]
  \]
  (Depending on how the domains are expanded, larger intervals could
  be obtained. Later in the paper, we explain how the above values
  were obtained.)
  Clearly, the modified rule is more informative, about the marks that
  yield a grade of B, than the rule shown in~\cref{ex:axp01}.
\end{example}

\begin{example}[Explanations with more expressive literals -- categorical features.]
  \label{ex:infxp02}
  For the decision list of~\cref{ex:runex02}, and instance
  $((\tn{Junior},\tn{Red}),1)$, the only AXp is $\fml{X}=\{1,2\}$.
  The purpose of this paper is to identify more general literals.
  For this example, instead of the rule from~\cref{ex:axp02}, a more
  expressive rule would be,
  \[
  \tn{IF} ~ A\:{\in}\:\{\tn{Junior},\tn{Senior}\} \land
  C\:{\in}\:\{\tn{Red},\tn{Blue},\tn{Green}\} ~ \tn{THEN} ~
  \kappa(\mbf{x})=1
  %
  \]
  As in the previous example, the modified rule is more informative
  than the rule shown in~\cref{ex:axp02}.
\end{example}

The use of generalized explanation literals is formalized in earlier
work~\cite{amgoud-ecsqaru21,amgoud-ijcai22,amgoud-ijar23}.
Initial approaches for computing explanations with more expressive
literals include~\cite{darwiche-corr20,iims-jair22,darwiche-corr23}.

\jnoteF{Discuss Darwiche's CoRR report~\cite{darwiche-corr20,darwiche-corr23}.}

\jnoteF{Discuss our own JAIR'22 paper~\cite{iims-jair22}.}

\section{Inflated Abductive Explanations} \label{sec:iaxp}

In order to account for more expressive literals in the definition of
abductive explanations, we consider an extended definition of AXp.

\subsection{Definition} \label{sec:ixp-def}
Given an AXp $\fml{X}\subseteq\fml{F}$, an inflated abductive
explanation (iAXp) is a tuple $(\fml{X},\mbb{X})$, with
$\fml{X}\subseteq\fml{F}$ is an AXp of the explanation problem
$\fml{E}$, and $\mbb{X}$ is a set of pairs $(j,\mbb{E}_j)$, one for
each  $j\in\fml{X}$, such that the following logic statement holds
true,
%
\begin{equation} \label{eq:axpxt}
  \forall(\mbf{x}\in\mbb{F}).%
  \left[\bigland_{j\in\fml{X}}(x_j\in\mbb{E}_j)\right]
  \limply(\kappa(\mbf{x})=c)
\end{equation}
where $v_j \in \mbb{E}_j, \forall j \in \fml{X}$, and where
$\mbb{E}_j$ is a maximal set such that~\eqref{eq:axpxt} holds.
(Concretely, for any $j\in\fml{X}$, and for any
$\mbb{I}_j\subseteq\mbb{D}_j\setminus\mbb{E}_j$, it is the case
that~\eqref{eq:axpxt} does not hold when $\mbb{E}_j$ is replaced with
$\mbb{E}_j\cup\mbb{I}_j$.)
Clearly,~\eqref{eq:axpxt} is a stronger statement than
Equation~(\ref{ex:axp01}), assuming at least one
$\mbb{E}_j\supset\{v_j\}$.

After computing one AXp $\fml{X}$, it is the case that either
$\mbb{E}_j=\{v_j\}$ (for a categorical feature) or
$\mbb{E}_j=[v_j,v_j]$ (for an ordinal feature). Our purpose is to find
ways of augmenting $\mbb{E}_j$ maximally.

\begin{example}
  For the monotonic classifier of~\cref{ex:runex01}, and the proposed
  inflated AXp of~\cref{ex:infxp01}, for feature 1, we can conclude
  that we get $\mbb{E}_1=[3.4,6.6]$.
\end{example}

\jnoteF{Def (first take): given an AXp $\fml{X}\subseteq\fml{F}$,
  and an order of the features in $\fml{X}$,
  $\iota:\fml{X}\to\{1,\ldots,|\fml{X}|\}$, an inflated explanation
  (defined on $\fml{X}$) is a rule of the form
  $\tn{IF}~\mathsf{COND}~\tn{THEN}~\mathsf{PRED}$, where each literal
  on feature $j\in\fml{X}$ in $\msf{COND}$, is of the form
  $x_j\in\mbb{E}_j$, and it is defined in terms of a maximal subset of
  $\mbb{E}_j\subseteq\mbb{D}_j$, given the literals defined for the
  other features in $\fml{X}$ with order smaller than the one of $j$.
}

Observe that the decision to start from an AXp aims at practical
efficiency.~\cref{alg:infxp} could be adapted to start from the set of
features $\fml{F}$ instead of $\fml{X}$. Any feature for which the
expansion includes its domain is removed from the AXp. The set of
inflated features for which the inflation did not yield their domain
represents one AXp.
%

\subsection{Computation of Inflated Explanations} \label{sec:ixp-alg}

\cref{alg:infxp} summarizes the algorithm for inflating a given AXp.
As shown, the algorithm picks some order for the features in the AXp
$\fml{X}$.
The features not included in the AXp $\fml{X}$ will not be analyzed,
i.e.\ we are only interested in inflating features that are not
already inflated.


\begin{algorithm}[t]
\hspace*{\algorithmicindent}
\textbf{Input}: {
  $\fml{E}=(\fml{M},(\mbf{v},c))$,
  AXp $\fml{X}\subseteq\fml{F}$,
  Precision $\delta$
}
\begin{algorithmic}[1]
  \Function{$\mathsf{InflateAXp}$}{$\fml{E},\fml{X}$} 
  \State{$\mbb{X}\gets\emptyset$}
  \Comment{$\mbb{X}$: Sets composing inflated explanation}
  \State{$\iota\gets\mathsf{PickSomeOrder}(\fml{X})$}
  \ForAll{$j\in\iota$}
  \If{$\msf{Categorical}(j)$}
  \State{$\mbb{E}_j\gets\{v_j\}$}
  \State{$\mbb{E}_j\gets\msf{InflateCategorical}(j,\mbb{E}_j,\fml{E},\fml{X})$}
  \Else
  \State{$\msf{inf}(j)=\msf{sup}(j)=v_j$}
  \State{$\mbb{E}_j\equiv[\msf{inf}(j),\msf{sup}(j)]$}
  \Comment{$\mbb{E}_j$ changes w/ $\msf{inf}(j)$,$\msf{sup}(j)$}
  \State{$\mbb{E}_j\gets\msf{InflateOrdinal}(j,\mbb{E}_j,\fml{E},\fml{X},\delta)$}
  \EndIf
  \State{$\mbb{X}\gets\mbb{X}\cup\{(j,\mbb{E}_j\})$}
  \EndFor
  \State{\Return $(\fml{X},\mbb{X})$}
  \EndFunction
\end{algorithmic}

  \caption{Computing inflated explanations}
  \label{alg:infxp}
\end{algorithm}

The procedure to inflate the values associated with a feature
$i\in\fml{X}$ depends on the type of the feature.
In the next sections, we detail how categorical and ordinal features
can be inflated.


\subsubsection{Categorical Features}

\cref{alg:iaxpcat} summarizes the algorithm for a given categorical
feature $j\in\fml{X}$.
%
%
%
%
%
Given an initial $\mbb{E}_j$, we traverse all the values in the domain
of the feature, with the exception of the values in $\mbb{E}_j$. For
each value, we check whether inflating $\mbb{E}_j$ with that value
maintains the sufficiency of prediction, i.e.~\cref{eq:axpxt} still
holds.
If sufficiency of prediction still holds, then the value is kept, and
the algorithm most to another value. Otherwise, the value is removed
from $\mbb{E}_j$, and another value will be considered.

\jnoteF{%
  Pick order $\iota$, for analyzing the features in $\fml{X}$.\\
  Let $j$ be a categorical) feature.
  \begin{enumerate}[nosep]
  \item Pick each value $u_j$ in $\mbb{D}_j\setminus\mbb{E}_j$;
  \item Add $u_j$ to $\mbb{E}_j$, and check whether \eqref{eq:axpxt}
    holds.
  \item If entailment is not preserved (i.e.\ \eqref{eq:axpxt} does
    not hold), then remove $u_j$ from $\mbb{E}_j$.
  \item Repeat the process for another value.
  \end{enumerate}
}

\begin{algorithm}[t]
\hspace*{\algorithmicindent}
\textbf{Input}: {Feature $j$,
  $\mbb{E}_j$,
  $\fml{E}=(\fml{M},(\mbf{v},c))$,
  and AXp $\fml{X}\subseteq\fml{F}$}
\begin{algorithmic}[1]
  \Function{$\mathsf{InflateCategorical}$}{$j,\mbb{E}_j,\fml{E},\fml{X}$}
  %
  \State{$\mbb{R}_j\gets\mbb{D}_j\setminus\mbb{E}_j$}
  \Comment{$\mbb{R}_j$: what remains of $\mbb{D}_j$}
  \State{$\eta\gets\mathsf{PickSomeOrder}(\mbb{R}_j)$}
  \ForAll{$u_{ji}\in\eta$}
  \State{$\mbb{E}_j\gets\mbb{E}_j\cup\{u_{ji}\}$}
  \Comment{Expand $\mbb{E}_j$, conditionally}
  \If{\eqref{eq:axpxt} does not hold}
  \State{$\mbb{E}_j\gets\mbb{E}_j\setminus\{u_{ji}\}$}
  \Comment{Revert expansion of $\mbb{E}_j$}
  \EndIf
  \EndFor
  \State{\Return $\mbb{E}_j$}
  \EndFunction
\end{algorithmic}

  \caption{Inflate categorical feature}
  \label{alg:iaxpcat}
\end{algorithm}

\begin{example}
  For the decision list of~\cref{ex:runex02}, and the instance
  $(\mbf{v},c)=((\tn{Junior},\tn{Red}),1)$, from~\cref{ex:axp02},
  we know that (the only) AXp is $\{1,2\}$.
  Let us pick the order of features $\iota=(1,2)$.
  For feature 1, the order of values is (for example)
  $(\tn{Adult},\tn{Senior})$, and initially
  $\mbb{E}_1=\{\tn{Junior}\}$. Clearly, we cannot consider the
  value $\tn{Adult}$ for feature 1, as this would change the
  prediction. In contrast, the value $\tn{Senior}$ can be added to
  $\mbb{E}_1$.
  With respect to feature 2, the order of values is (for example)
  $(\tn{Blue},\tn{Green},\tn{Silver},\tn{Black},\tn{White})$,
  and initially $\mbb{E}_2=\{\tn{Red}\}$.
  Clearly, by inspection of the decision list, we can conclude that,
  for a non-Adult, any of the colors Red, Blue, Green and Black will
  yield prediction 1.
  Hence,
  $\mbb{E}_2$ can be inflated from $\{\tn{Red}\}$ to
  $\{\tn{Red},\tn{Blue},\tn{Green},\tn{Black}\}$, as these values do
  not change the sufficiency for the prediction.
  As a result, the inflated explanation denotes the rule,
  \begin{align}
    \tn{IF} ~ & A\:{\in}\:\{\tn{Junior},\tn{Senior}\} \land \nonumber\\
    & C\:{\in}\:\{\tn{Red},\tn{Blue},\tn{Green},\tn{Black}\} ~ \tn{THEN} ~
  \kappa(\mbf{x})=1 \nonumber 
  \end{align}
  Hence, from the decision list, and given the list, it becomes clear
  that a driver does not pose a risk of accident if he/she is an
  Adult, or otherwise drive a car colored Silver or White.
\end{example}

\subsubsection{Ordinal Features}

\begin{algorithm}[t]
\hspace*{\algorithmicindent}
\textbf{Input}: {%
  Feat.~$j$,
  $\mbb{E}_j$,
  $\fml{E}=(\fml{M},(\mbf{v},c))$,
  AXp $\fml{X}\subseteq\fml{F}$,
  Precision $\delta$}
\begin{algorithmic}[1]
  \Function{$\mathsf{InflateOrdinal}$}{$j,\mbb{E}_j,\fml{E},\fml{X},\delta$} 
  %
  \If{$\mathsf{UnconstrainedSup}(j,\fml{E},\fml{X})$}
  \State{$\msf{sup}(j)\gets\mu(j)$}
  \If{$\mu(j)={+}\infty$}
  \State{$\mbb{E}_j\equiv[\msf{inf}(j),{+}\infty)$}
  \EndIf
  \Else
  \State{$\msf{sup}(j)\gets\mathsf{ExpandSup}(j,\fml{E},\fml{X},\delta)$}
  \EndIf
  \If{$\mathsf{UnconstrainedInf}(j,\fml{E},\fml{X})$}
  \State{$\msf{inf}(j)\gets\lambda(j)$}
  \If{$\lambda(j)={-}\infty$}
  \State{$\mbb{E}_j\equiv({-}\infty,\msf{sup}(j)]$}
  \EndIf
  \Else
  \State{$\msf{inf}(j)\gets\mathsf{ExpandInf}(j,\fml{E},\fml{X},\delta)$}
  \EndIf
  \State{\Return $\mbb{E}_j$}
  \EndFunction
\end{algorithmic}

  \caption{Inflate ordinal feature}
  \label{alg:iaxpord}
\end{algorithm}

\cref{alg:iaxpord} summarizes the algorithm for inflating a given
ordinal feature $j\in\fml{X}$.
%
%
%
%
With the purpose of assessing the largest and smallest possible values
for feature $j$, we initially check whether the geature can take its
upper bound $\mu(j)$ (or can be unbounded when $\mu(j)={+}\infty$),
and later check whether the feature can take its lower
bound. (Evidently, no feature $j$ in $\fml{X}$ can be allowed to take
any value between its lower and upper bounds; since the feature $j$ is
included in the AXp, then it cannot be allowed to take any value in
its domain.)
If the feature is unconstrained for either its largest or smallest
value, then $\mbb{E}_j$ is updated accordingly (with the case of
${+}\infty$/${-}\infty$ being handled differently). Otherwise, we seek
to inflate $\mbb{E}_j$, either by increasing values or by decreasing
values.
The search for increasing values is illustrated
by~\cref{alg:expandSup}, and the search for decreasing values is
illustrated by~\cref{alg:expandInf}.
For simplicity, the algorithms implement a (simple) linear search,
using only the value of precision $\delta$ for approximating the
solution. If $\delta$ is small this is inefficient.
A simple improvement consists in using two positive values, $\beta$
and $\delta$, where (the larger) $\beta$ is used for a rough
approximation of the solution, and (the smaller) $\delta$ is then used
to refine the coarser approximation obtained with $\beta$.
Moreover, assuming the upper (or lower) bound is finite, then standard
binary search can be used. If the upper (or lower) bound is infinite,
and not constrained, then one can use exponential (binary)
search%
\footnote{%
Exponential binary search is a common algorithm for finding a value
when the domain is unbounded~\cite{bentley-ipl76}, which is guaranteed
to terminate if one knows that the target value exists, and given some
value of precision. 
},
with a subsequent binary search step to zoom in on the largest (or
smallest) value of the feature's values that guarantee sufficiency for
the prediction.

\begin{algorithm}[t]
\hspace*{\algorithmicindent}
\textbf{Input}: {%
  Feature $j$,
  $\fml{E}=(\fml{M},(\mbf{v},c))$,
  AXp $\fml{X}\subseteq\fml{F}$,
  Precision $\delta$}
\begin{algorithmic}[1]
  \Function{$\mathsf{expandSup}$}{$j,\fml{E},\fml{X},\delta$}
  \While{\tbf{true}}
  \State{$\msf{sup}(j)\gets\msf{sup}(j) + \delta$}
  \If{\eqref{eq:axpxt} does not hold}
  \State{$\msf{sup}(j)\gets\msf{sup}(j) - \delta$}
  \State{\Return $\msf{sup}(j)$}
  \EndIf
  \EndWhile
  \EndFunction
\end{algorithmic}

  \caption{Expanding the supremum with linear search}
  \label{alg:expandSup}
\end{algorithm}

\begin{algorithm}[t]
\hspace*{\algorithmicindent}
\textbf{Input}: {%
  Feature $j$,
  $\fml{E}=(\fml{M},(\mbf{v},c))$,
  AXp $\fml{X}\subseteq\fml{F}$,
  Precision $\delta$}
\begin{algorithmic}[1]
  \Function{$\mathsf{expandInf}$}{$j,\fml{E},\fml{X},\delta$}
  \While{\tbf{true}}
  \State{$\msf{inf}(j)\gets\msf{inf}(j) - \delta$}
  \If{\eqref{eq:axpxt} does not hold}
  \State{$\msf{inf}(j)\gets\msf{inf}(j) + \delta$}
  \State{\Return $\msf{inf}(j)$}
  \EndIf
  \EndWhile
  \EndFunction
\end{algorithmic}

  \caption{Expanding the infimum with linear search}
  \label{alg:expandInf}
\end{algorithm}

\begin{example}
For the example monotonic classifier (see~\cref{ex:runex01}), with
instance $((5,10,5,8),B)$, the computed AXp is $\{1,2,4\}$. Let
$\delta=0.2$.
We illustrate the execution of~\cref{alg:iaxpord},

in the concrete
case where we increase the maximum value that $Q$ (i.e.\ feature 1)
can take such that the prediction does not change.
Since the classifier is monotonically increasing, we can fix $H$ to
10. The value of $X$ is set to 10 and $R$ is set to 8.
Clearly, $Q$ cannot take value 10; otherwise the prediction would
become $A$.
By iteratively increasing the largest possible value for $Q$, with
increments of $\delta$ (see~\cref{alg:expandSup}), we conclude that
$Q$ can increase up to 6.6 while ensuring that the prediction does not
change to $A$. If $Q$ were assigned value 6.8 (the next $\delta$
increment), the prediction would change to $A$.
In terms of the smallest value that can be assigned to $Q$, we
assign $H$ to 0.
In this case, we conclude that $Q$ can take the (lower bound) value of
0, because $R=8$.
%
%
%
%
%
Hence, feature $Q$ can take values in the range $[0,6.6]$.
%
%
Finally, in the case of $R$, its value cannot get to 9, since
otherwise the prediction would become $A$. Hence, the value of $R$
before considering 9 is 8.8 (see~\cref{ex:infxp01}).
In terms of the smallest possible value for $R$, it is clear that its
value cannot be less than 7, as this would serve to change the
prediction.
Hence, feature $R$ can take values in the range $[7,8.8]$.
%
Given the above, the rule associated with the inflated AXp is:
\[
\tn{IF}~[{Q}\in[0,6.6]\land{R}\in[7,8.8]]~\tn{THEN}~[\kappa(\mbf{x})=B]
\]
\end{example}

\subsubsection{Ordinal Features and Tree-based models}

Tree-based machine learning models such as decision trees, random forest,
and boosted trees allow a simpler treatment of ordinal features $j$ since
the trees will only compare to a finite set of feature values $V_j \subset
\mathbb{D}_j$, we assume with comparison $x_j \geq d, d \in V_j$. Let $[d_1, \ldots,
  d_m]$ be the $V_j$ is sorted order.
We can construct disjoint intervals, given by
$I_1 = [\min(\mathbb{D}_j), d_1)$, $I_2 = [d_1, d_2)$, \ldots, $I_{m+1} =
    [d_m,\max(\mathbb{D}_j)]$.
By construction no two values in any interval can be treated differently by
the tree-base model, hence we can use these intervals as a finite
categorical representation of the feature~$j$.

\subsection{Complexity of Inflated Explanations}
As can be concluded from the algorithms described in the previous
section, sufficiency for prediction is checked using~\cref{eq:axpxt},
which mimics the oracle call for finding the AXp.
Nevertheless, inflated explanations require a number of calls
to~\cref{eq:axpxt} that grows with the number of values in the
features domains (for categorical features), or the computed supremum
(or the infimum) divided by $\delta$, for categorical features when
linear search is used. (The analysis for (unbounded) binary search is
beyond the goals of the paper.)

\section{Inflated Contrastive Explanations \& Duality}
\label{sec:icxp}

This section investigates the differences that must be accounted for
in the case of contrastive explanations.

\subsection{Plain Inflated CXp's \& MHS Duality}

In contrast with AXp's, in the case of CXp's inflation takes place in
the features \emph{not} in the explanation, i.e.\ in the features whose
value remains fixed.

As with AXp's, we propose a modification to the definition of CXp, as
follows:
\begin{equation} \label{eq:cxpxt}
  \exists(\mbf{x}\in\mbb{F}).%
  \left[\bigland_{j\in\fml{F}\setminus\fml{Y}}(x_j\in\mbb{E}_j)\right]%
  \land(\kappa(\mbf{x})\not=c)
\end{equation}

Minimal hitting set duality between inflated AXp's and inflated CXp's
is immediate and follows from the duality result relating AXp's and
CXp's~\cite{inams-aiia20}.
%
%
To motivate the argument, let us define the following sets:
\begin{align}
  \mbb{A}(\fml{E})&=\{\fml{X}\subseteq\fml{F}\,|\,\axp(\fml{X})\}\\
  \mbb{C}(\fml{E})&=\{\fml{X}\subseteq\fml{F}\,|\,\cxp(\fml{Y})\}
\end{align}
Given the discussion above, the following result follows:
\begin{proposition} \label{prop:unchange}
  $\mbb{A}(\fml{E})$ and $\mbb{C}(\fml{E})$ remain unchanged for
  inflated explanations.
\end{proposition}
Given~\cref{prop:unchange} and from earlier work~\cite{inams-aiia20},
it follows that inflated AXp's and minimal hitting sets of the set of
CXp's and vice-versa.
%

Nevertheless, and in contrast with~\eqref{eq:axpxt},~\eqref{eq:cxpxt}
is a weaker definition that the original definition of CXp.
The next section proposes a stronger definition of CXp.

\subsection{Generalized CXp's \& Extended  Duality}

\jnoteF{Goals:\\
  Basic set-based duality between AXp's and CXp's.\\
  Additional comments on duality given values used.
}

Due to the use of sets $\mbb{E}_j\subset\mbb{D}_j$ instead of concrete
feature-values $v_j\in\mbb{D}_j$, the straightforward definition of
inflated contrastive explanations above provides a weaker explanation,
than the uninflated definition~\eqref{eq:cxp}.
Nevertheless, it allows us to establish a simple minimal hitting set
duality between inflated AXp's and inflated CXp's by building directly
on~\cite{inams-aiia20}.
In particular, each inflated CXp is a minimal hitting set of all the
inflated AXp's and vice versa, in the sense that given the set of all
inflated AXp's (resp. inflated CXp's), an uninflated CXp (resp.
uninflated AXp) can be constructed by this duality and then inflated
afterwards.
Note that this \emph{set-wise} duality requires us to first
reconstruct uninflated ``versions'' of the dual explanations and only
then inflate them.
We can define a stronger form of inflated contrastive explanation as
follows, which will enable us to construct inflated CXp's (resp.
inflated AXp's) directly from inflated AXp's (resp. inflated CXp's).
Namely, given an instance $(\mbf{v}, c)$ and an inflated CXp is a pair
$(\fml{Y}, \mbb{Y})$ s.t\ $\mbb{Y}$ is a set of pairs $(j, \mbb{G}_j)$
for each feature $j\in\fml{Y}$, such that the following holds:
\begin{equation} \label{eq:cxpxt2}
  \exists(\mbf{x}\in\mbb{F}).%
  \left[\bigland_{j\in\fml{F}\setminus\fml{Y}}(x_j = v_j)
\land \bigland_{j\in\fml{Y}}(x_j \in \mathbb{G}_j)\right]%
  \land(\kappa(\mbf{x})\not=c)
\end{equation}

%
Note that we can assume $v_j \not\in \mathbb{G}_j, j \in \fml{Y}$
since otherwise we could eliminate $j$ from $\fml{Y}$ and have a
tighter contrastive explanation.
Also, observe that in contrast to~\eqref{eq:cxp},
Equation~\eqref{eq:cxpxt2} considers each of the features
$j\in\fml{Y}$ to belong to some set $\mbb{G}_j\subset\mbb{D}_j$ rather
than assuming them to be free.

The algorithms proposed in~\cref{sec:iaxp} can be adapted for
inflating CXp's by checking whether~\eqref{eq:cxpxt2} holds instead
of~\eqref{eq:axpxt}.
An immediate observation here is that these algorithms would need to
be updated such that given a feature $j\in\fml{Y}$, we need to
\emph{shrink} the set of allowed values starting from $\mbb{D}_j$
rather than inflating it starting from $\{v_j\}$, which was the case
for inflated AXp's.


In order to facilitate the duality relationship between inflated AXp's
and inflated CXp's, observe that given an instance $(\mbf{v}, c)$, an
inflated AXp $(\fml{X}, \mbb{X})$, as defined in~\eqref{eq:axpxt}, can
be equivalently reformulated as follows:
\begin{equation} \label{eq:axpxt-ex}
  \forall(\mbf{x}\in\mbb{F}).%
  \left[\bigland_{j\in\fml{X}}(x_j\in\mbb{E}_j) \wedge \bigland_{j \in
      \fml{F} \setminus \fml{X}} x_j \in \mathbb{D}_j\right]%
  \limply(\kappa(\mbf{x})=c)
\end{equation}
In other words, simply let $\mathbb{E}_j = \mathbb{D}_j$ for all $j
\not\in \fml{X}$.
Similarly, given an instance $(\mbf{v}, c)$, an inflated CXp
$(\fml{Y}, \mbb{Y})$, s.t.\ $\mbb{Y}$ is a set of pairs $(j,
\mbb{G}_j)$, can be reformulated such as the following holds:
\begin{equation} \label{eq:cxpxt-ex}
  \exists(\mbf{x}\in\mbb{F}).%
  \left[\bigland_{j\in\fml{F}\setminus\fml{Y}}(x_j \in \{ v_j\})
\land \bigland_{j\in\fml{Y}}(x_j \in \mathbb{G}_j)\right]%
  \land(\kappa(\mbf{x})\not=c)
\end{equation}
In other words, $\mathbb{G}_j = \{v_j\}$ for all $j \not\in \fml{Y}$.


\begin{proposition}
  Given an explanation problem $\fml{E}$, let $\mbb{A}'(\fml{E})$
  denote the set of all iAXps $(\fml{X},\mbb{X})$ while
  $\mbb{C}'(\fml{E})$ denote the set of all iCXps $(\fml{Y},\mbb{Y})$.
  Then each iAXp $(\fml{X},\mbb{X})\in\mbb{A}'(\fml{E})$ minimally
  ``hits'' each iCXp $(\fml{Y},\mbb{Y})\in\mbb{C}'(\fml{E})$ s.t. if
  feature $j\in\fml{F}$ is selected to ``hit'' iAXp
  $(\fml{X},\mbb{X})$ then $\mbb{G}_j\cap\mbb{E}_j=\emptyset$, and
  vice versa.
  \begin{proof}
    Suppose given an iAXp $(\fml{X},\mbb{X})$ and a iCXp
    $(\fml{Y},\mbb{Y})$, feature $j\in\fml{F}$ is selected to hit the
    CXp such that $\mbb{G}_j\cap\mbb{E}_j\neq\emptyset$.
    Then we have a contradiction.
    Indeed, we can extract an instance $\mathbf{v'} = (v'_1, \ldots,
    v'_m)$ s.t. $v_j'\in\mbb{G}_j\cap\mbb{E}_j$ satisfying Equation
    (\ref{eq:cxpxt-ex}),
    with class $\kappa(\mathbf{v'}) \neq c$, which by definition
    ~(\ref{eq:axpxt-ex}) violates the iAXp condition since $v'_j \in \mathbb{G}_j
    \cap \mathbb{E}_j$.
    %
    %
  \end{proof} \end{proposition}

Given the above proposition, we can construct inflated CXp's from
inflated AXp's as follows.
Given a complete set $\mbb{A}'(\fml{E})$ of inflated AXp's, we select
one feature $\theta(\fml{X}) \in \fml{X}$ in the features for
explanation $(\fml{X},\mbb{X})$ and construct an inflated CXp defined
by subset of features $\fml{Y} = \{ \theta(\fml{X}) ~|~
(\fml{X},\mbb{X}) \in \mbb{A}'(\fml{E}) \}$ and $\mathbb{G}_j =
\bigcap_{(\fml{X,\mbb{X}}) \in \mbb{A}'(\fml{E}), \theta(\fml{X}) = j}
(\mathbb{D}_j \setminus \mathbb{E}_j^\fml{X})$.
Similarly, given a complete set $\mbb{C}'(\fml{E})$ of inflated CXp's,
we can construct an inflated AXp, by selecting one feature
$\phi(\fml{Y}) \in \fml{Y}$ from each CXp $(\fml{Y},\mbb{Y})$, and
defining $\fml{X} = \{ \phi(\fml{Y}) ~|~ \fml{Y} \in \mbb{C}'(\fml{E})
\}$ and $\mathbb{E}_j =  \bigcap_{(\fml{Y,\mbb{Y}}) \in
\mbb{C}'(\fml{E}), \phi(\fml{Y}) = j} (\mathbb{D}_j \setminus
\mathbb{G}_j^\fml{Y})$.
Note that the key difficulty here is to organize efficient
\emph{minimal} selection of features used to ``hit'' the given set of
explanations (either $\mbb{A}'(\fml{E})$ or $\mbb{C}'(\fml{E})$).
In various settings of minimal hitting set
duality~\cite{inams-aiia20}, this is typically done by invoking a
modern mixed-integer linear programming (MILP) or maximum
satisfiability (MaxSAT) solver.
Similar ideas can be applied in this case as well.


\section{Experiments} \label{sec:res}
\sisetup{parse-numbers=false,detect-all,mode=text}
\setlength{\tabcolsep}{5pt}
\let\lpr\undefined
\let\rpr\undefined
\newcommand{\lpr}{(}
\newcommand{\rpr}{)}

\begin{table*}[ht]
\centering
\resizebox{0.85\textwidth}{!}{
  \begin{tabular}{l>{\lpr}S[table-format=2.0,table-space-text-pre=\lpr]S[table-format=3.0,table-space-text-post=\rpr]<{\rpr}
  S[table-format=2]S[table-format=5.0]S[table-format=2.0]
  S[table-format=2.1]S[table-format=1.2]S[table-format=3.0]
  S[table-format=4.0]S[table-format=3.1]S[table-format=3.2]}
\toprule[1.2pt]
\multirow{2}{*}{\bf Dataset} & \multicolumn{2}{c}{\multirow{2}{*}{\bf (m,~K)}}  & \multicolumn{3}{c}{\bf RF} &
\multicolumn{2}{c}{\bf AXp }  &   \multicolumn{4}{c}{\bf Inflated AXp}  \\
  \cmidrule[0.8pt](lr{.75em}){4-6}
  \cmidrule[0.8pt](lr{.75em}){7-8}
  \cmidrule[0.8pt](lr{.75em}){9-12}
& \multicolumn{2}{c}{} & {\bf D}  & {\bf \#N} & {\bf \%A} & {\bf Len }  & {\bf Time } & {\bf m } & {\bf M } & {\bf avg } &  {\bf Time } \\
\toprule[1.2pt]

adult  &  12  &  2  &  8  &  21526  &  76  &  5.6  &  0.19  &  0  &  26  &  7.0  &  0.40 \\
ann-thyroid  &  21  &  3  &  4  &  2192  &  96  &  1.7  &  0.20  &  6  &  129  &  15.8  &  0.41 \\
appendicitis  &  7  &  2  &  6  &  1920  &  90  &  3.7  &  0.05  &  13  &  140  &  54.2  &  0.33 \\
banknote  &  4  &  2  &  5  &  2772  &  98  &  2.2  &  0.05  &  16  &  351  &  156.6  &  0.22 \\
biodegradation  &  41  &  2  &  5  &  4420  &  88  &  16.7  &  0.33  &  47  &  611  &  273.1  &  6.46 \\
ecoli  &  7  &  5  &  5  &  3860  &  90  &  3.6  &  0.20  &  48  &  242  &  128.4  &  1.39 \\
german  &  21  &  2  &  8  &  13332  &  74  &  12.3  &  0.36  &  7  &  435  &  149.0  &  4.20 \\
glass2  &  9  &  2  &  6  &  2966  &  84  &  4.6  &  0.07  &  6  &  263  &  92.9  &  0.41 \\
heart-c  &  13  &  2  &  5  &  3910  &  83  &  5.5  &  0.07  &  9  &  207  &  89.1  &  0.23 \\
ionosphere  &  34  &  2  &  5  &  2096  &  85  &  21.5  &  0.06  &  16  &  352  &  193.5  &  0.27 \\
iris  &  4  &  3  &  6  &  1446  &  93  &  2.2  &  0.11  &  14  &  48  &  20.7  &  0.15 \\
karhunen  &  64  &  10  &  5  &  6198  &  92  &  34.8  &  3.07  &  215  &  650  &  392.4  &  17.14 \\
lending  &  9  &  2  &  5  &  5370  &  75  &  2.2  &  0.10  &  0  &  6  &  1.7  &  0.14 \\
magic  &  10  &  2  &  6  &  9840  &  85  &  6.3  &  0.25  &  81  &  2041  &  831.7  &  6.84 \\
mofn-3-7-10  &  10  &  2  &  6  &  8776  &  91  &  3.0  &  0.06  &  0  &  0  &  0.0  &  0.06 \\
mushroom  &  22  &  2  &  4  &  1930  &  97  &  9.2  &  0.07  &  1  &  42  &  14.5  &  0.15 \\
new-thyroid  &  5  &  3  &  5  &  1766  &  100  &  2.9  &  0.12  &  9  &  108  &  60.2  &  0.40 \\
pendigits  &  16  &  10  &  6  &  12004  &  96  &  9.8  &  1.07  &  89  &  569  &  281.0  &  12.50 \\
phoneme  &  5  &  2  &  6  &  8962  &  83  &  3.0  &  0.18  &  151  &  1765  &  582.5  &  2.62 \\
promoters  &  57  &  2  &  3  &  1214  &  90  &  25.0  &  0.05  &  13  &  38  &  23.0  &  0.12 \\
recidivism  &  15  &  2  &  10  &  75204  &  61  &  7.9  &  0.48  &  0  &  6  &  2.2  &  0.53 \\
ring  &  20  &  2  &  6  &  6188  &  90  &  9.6  &  0.35  &  14  &  1639  &  720.5  &  1.35 \\
segmentation  &  19  &  7  &  4  &  1966  &  90  &  8.1  &  0.39  &  19  &  149  &  59.7  &  1.99 \\
shuttle  &  9  &  7  &  3  &  1460  &  99  &  2.3  &  0.22  &  7  &  50  &  19.8  &  0.42 \\
sonar  &  60  &  2  &  5  &  2614  &  88  &  35.9  &  0.11  &  152  &  327  &  242.8  &  4.10 \\
soybean  &  35  &  18  &  5  &  4460  &  87  &  15.6  &  2.05  &  4  &  35  &  13.1  &  3.39 \\
spambase  &  57  &  2  &  5  &  4614  &  92  &  18.9  &  0.15  &  57  &  516  &  211.6  &  0.94 \\
spectf  &  44  &  2  &  5  &  2306  &  87  &  19.7  &  0.10  &  11  &  221  &  108.9  &  2.11 \\
texture  &  40  &  11  &  5  &  5724  &  89  &  22.8  &  1.06  &  70  &  834  &  338.4  &  13.19 \\
threeOf9  &  9  &  2  &  3  &  920  &  100  &  1.0  &  0.02  &  0  &  0  &  0.0  &  0.02 \\
twonorm  &  20  &  2  &  3  &  1500  &  92  &  10.4  &  0.05  &  17  &  248  &  99.2  &  0.15 \\
vowel  &  13  &  11  &  6  &  10176  &  91  &  7.7  &  1.28  &  70  &  862  &  401.5  &  28.05 \\
waveform-21  &  21  &  3  &  5  &  6238  &  84  &  9.6  &  0.50  &  81  &  1026  &  373.6  &  6.28 \\
waveform-40  &  40  &  3  &  5  &  6232  &  84  &  14.5  &  0.90  &  73  &  975  &  365.0  &  8.21 \\
wdbc  &  30  &  2  &  4  &  2028  &  94  &  11.9  &  0.08  &  56  &  274  &  119.8  &  0.45 \\
wine-recog  &  13  &  3  &  3  &  1188  &  97  &  4.5  &  0.15  &  10  &  112  &  39.0  &  0.35 \\
wpbc  &  33  &  2  &  5  &  2432  &  76  &  19.9  &  0.80  &  89  &  429  &  211.2  &  144.48 \\
xd6  &  9  &  2  &  6  &  8288  &  100  &  3.0  &  0.07  &  0  &  0  &  0.0  &  0.06 \\

\bottomrule[1.2pt]
\end{tabular}
}
\caption{%
  Detailed performance evaluation of delivering inflated AXp's for RFs.
The table shows results for 38 datasets, which contain categorical and 
ordinal  data.
Columns {\bf m} and {\bf K} report, respectively, the
number of features and classes in the dataset.
Columns {\bf D}, {\bf\#N} and {\bf\%A} show, respectively, each tree's
max.~depth, total number of nodes and test accuracy of an RF classifier.
Column {\bf Len} reports the average explanation length
(i.e. average number of features contained in the explanations).
Column {\bf Time}   shows the average  runtime  for extracting
an explanation.
Column  {\bf avg} (resp.\ m and M) reports the average number of 
values/intervals (for categorical /continuous domain) computed 
in the expansion of $\mbb{E}_j$ for the inflated AXp. 
}
\label{tab:RFs-sat}
\end{table*}

This section presents a summary of empirical assessment
of computing inflated abductive explanations for the case study of RF classifiers
trained on some of the widely studied datasets.

\paragraph{Experimental setup.}
The experiments are conducted on a MacBook Pro with a Dual-Core Intel
Core~i5 2.3GHz CPU with 8GByte RAM running macOS Ventura.
The results reported do not impose any time or memory limit.

\paragraph{Benchmarks.}
The assessment is performed on a selection of 38 publicly available
datasets, which originate from UCI Machine Learning Repository
\cite{uci} and Penn Machine Learning Benchmarks~\cite{Olson2017PMLB}.
The benchmarks comprise  binary and multidimensional classification datasets
and include ordinal and categorical datasets.
(Categorical features are encoded into OHE data and handled as
a group of attributes representing their original features when computing
the explanations.)
The number of classes in the benchmark suite varies from 2 to 18.
The number of features  varies from 4 to 64
with the average being 22.5 (resp.~5131.09).
When training RF classifiers for the selected datasets, we used 80\% of
the dataset instances (20\% used for test data). For assessing
explanation tools, we randomly picked fractions of the dataset,
depending on the dataset size.
The scikit-learn toolkit~\cite{scikitlearn} was used to train the models.
The number of trees in each learned model is set to 100, while tree depth
varies between 3 and 10.
As a result, the accuracy of the trained models varies between 61\%
to 100\%.
Besides, our formal explainers are set to compute a single  AXp
and then apply the inflation method, per data instance from
the selected set of instances and 200 samples are randomly
to be tested for each dataset.

\paragraph{Prototypes implementation.}
We developed a reasoner for RFs as Python script\footnote{%
The Python code is publicly available on 
\url{https://github.com/yizza91/RFxpl}}. The script
implements the SAT-based encoding of RFs proposed
in~\cite{ims-ijcai21}, and the outlined algorithms
(\cref{alg:infxp,alg:iaxpcat,alg:iaxpord,alg:expandSup,alg:expandInf})
in order to compute inflated AXp's.
Also, the prototype can serve to compute first an AXp, if this one was
not provided
in the inputs, and then generate an inflated  explanation.
Furthermore, PySAT~\cite{imms-sat18} is used to generate the CNF formulas
and cardinality constraints in a clausal form to encode the classifier. Afterwards,
PySAT is used to  instrument incremental
calls to the Glucose3~\cite{audemard2018glucose} SAT solver when computing explanations.

\paragraph{Results.}
\autoref{tab:RFs-sat} summarizes the results of computing inflated AXp's for
RFs on the selected datasets.
As can be observed from the results, and with three exceptions, our method
succeeds in expanding  all individual set $\mbb{E}_i$ of features involved in
the AXp.
%
Moreover, we observe that for 20 out of 38 datasets, the average increase
in sub-domain $\mbb{E}_i$  varies between 100 and 720, and
for 13 over 38 datasets this number
varies between 13 to 99.
In terms of performance, the results clearly demonstrate that our approach
scales to (realistically) large data and large tree ensembles  considered in
the assessment.
It is plain to see that for most datasets the proposed method takes a few seconds on
average to deliver an inflated AXp, thus the average
(resp.\ minimum and maximum)
runtime for all datasets is 7.11 seconds (resp. 0.02 and 144.48 seconds).
Even though a few outliers were observed in continuous data
where the number of splits (intervals) generated by the trees is fairly large,
this does not contrast the effectiveness of our technique since the largest
running time that could be registered is less than ~2.30 minutes.
As results, our extensive evaluation performed on a large range
of real world data and RFs of large sizes, allows us to conclude that our solution
is effective in practice to produce more expressive explanations
than standard AXp's and 
more importantly in a short time.

\section{Conclusions} \label{sec:conc}

One limitation of logic-based explanations, either abductive or
contrastive explanations, is that these are based on fairly restricted
literals, of the form $x_i=v_i$.
This paper formalizes the concept of \emph{inflated explanation},
which can be considered either in the case of abductive or contrastive
explanations. Furthermore, the paper proposes algorithms for the
rigorous computation of inflated explanations, and demonstrates the
existence of minimal hitting set duality between inflated abductive
and inflated contrastive explanations.
The experimental results validate the practical interest of computing
inflated explanations.


\newtoggle{mkbbl}

\settoggle{mkbbl}{false}

\iftoggle{mkbbl}{
  \bibliographystyle{named}
  \bibliography{team,refs,nfxai}
}{

}

\end{document}